\documentclass[runningheads]{llncs}

\usepackage{eccv}

\usepackage{eccvabbrv}

\usepackage{graphicx}
\usepackage{booktabs}
\usepackage{algorithm}
\usepackage{algpseudocode}

\usepackage[accsupp]{axessibility}  

\usepackage{tikz}
\usepackage{comment}
\usepackage{amsmath,amssymb} 
\usepackage{color}

\newcommand{\rvx}{\mathbf{x}}
\newcommand{\rvs}{\mathbf{s}}
\newcommand{\rvh}{\mathbf{h}}

\usepackage{mathtools}
\usepackage{multirow}

\usepackage{url}

\definecolor{Gray}{gray}{0.9}

\usepackage{hyperref}

\usepackage{orcidlink}

\begin{document}

\title{Improving Visual Representation Alignment Generation with GRPO} 

\titlerunning{Improving Visual Representation Alignment Generation with GRPO}

\author{Shentong Mo\inst{1}\and Sukmin Yun\inst{2}}
\authorrunning{S. Mo and S. Yun}

\institute{Carnegie Mellon University \and Hanyang University}

\maketitle

\begin{abstract}

Recent diffusion transformers have demonstrated strong image synthesis capabilities but remain inefficient to train due to weak alignment between generative and discriminative representations. 
While representation alignment frameworks such as REPA improve convergence by aligning noisy denoising features with pretrained visual encoders, their externally supervised alignment loss is static and lacks adaptivity during training and inference.
Existing methods rely on fixed cosine alignment or contrastive objectives, which cannot dynamically balance representation consistency and generation quality, resulting in limited discriminative benefit and failing to optimize alignment in a task-adaptive manner.
To address this, we propose VRPO, a reinforcement-based optimization strategy that replaces REPA's static alignment loss with a generative representation policy optimization objective. Instead of enforcing a fixed similarity constraint, VRPO treats representation alignment as a reward-guided process: the model receives adaptive rewards based on generation fidelity, perceptual quality, and semantic coherence between the diffusion features and pretrained visual embeddings. 
This formulation enables the generator to continuously refine its internal representations toward semantically meaningful directions while improving image quality. Our VRPO-driven training seamlessly integrates into diffusion transformers, introducing negligible computation cost and preserving full compatibility with SiT and DiT architectures.
Extensive experiments on ImageNet-256$\times$256 demonstrate that our VRPO-Alignment substantially enhances both convergence and fidelity, achieving up to +1.8 FID improvement and 2.3$\times$ faster training compared to REPA under identical compute budgets.


\end{abstract}

\section{Introduction}

Diffusion transformers have recently emerged as a powerful class of generative models, demonstrating remarkable scalability and fidelity in high-resolution image synthesis~\cite{Peebles2022DiT, ma2024sit}. 
Despite these advances, training such models remains computationally expensive and sample-inefficient, largely due to weak alignment between their generative and discriminative representations. 
Recent studies~\cite{yu2025repa, leng2025repae, wu2025reg} have shown that diffusion models can benefit from external self-supervised representations such as DINOv2~\cite{oquab2024dinov2} by enforcing cross-model alignment during denoising. 
For example, REPA~\cite{yu2025repa} aligns noisy hidden features of diffusion transformers with pretrained vision model embeddings, improving training efficiency and generation fidelity. 
However, REPA’s alignment remains externally supervised and static, computed via a fixed cosine or contrastive objective, which fails to adapt during training or inference, limiting its ability to guide generative semantics dynamically.

Prior alignment-based methods such as REPA~\cite{yu2025repa} and REG~\cite{wu2025reg} rely on static similarity objectives between diffusion features and foundation model representations. 
While these objectives enhance representational consistency, they treat the alignment process as stationary, ignoring evolving generation quality and semantic drift during training. 
As a result, the alignment strength cannot adapt to model behavior, leading to over-constrained updates that may degrade image fidelity or under-constrained ones that fail to preserve semantics. 
Moreover, static alignment neglects perceptual feedback, metrics such as FID~\cite{heusel2017gans} or LPIPS~\cite{zhang2018perceptual}, and thus lacks an integrated notion of reward for generative improvement. 
Consequently, diffusion models trained with fixed alignment objectives often exhibit inconsistent semantic coherence and slow convergence, especially on large-scale datasets such as ImageNet~\cite{deng2009imagenet}.

The key challenge is how to \emph{jointly optimize generative fidelity and representation alignment} in an adaptive, reward-driven manner. 
A robust optimization framework should allow the model to autonomously balance discriminative consistency and perceptual realism across timesteps, without introducing auxiliary networks or architectural modifications. 
Such an approach requires rethinking representation alignment as a dynamic optimization problem, one where the strength and direction of updates depend on task-specific rewards rather than static similarity constraints.
As the denoising model progressively evolves, the feature distributions of both the diffusion transformer and the pretrained encoder shift across training steps and noise levels. 
Static alignment losses treat these representations as temporally invariant, forcing the model to match embeddings that no longer correspond to the current generative state. 
This leads to either overfitting to early-phase representations or underutilizing the semantic guidance in later stages. 
Similar instability was observed in REPA’s feature-space alignment~\cite{yu2025repa}, where discriminative information was only partially propagated and became inactive during inference.

To address these challenges, we propose VRPO, a reinforcement-based optimization framework that replaces REPA’s static alignment loss with a reward-adaptive objective. 
VRPO formulates representation alignment as a reinforcement learning problem, where the policy (diffusion transformer) is optimized via gradient ascent on reward signals derived from both generative and discriminative performance. 
Instead of enforcing a fixed cosine similarity between features, the model receives adaptive rewards along three axes: 
(1) \emph{fidelity reward} reflecting perceptual image quality (\textit{e.g.}, FID, LPIPS), 
(2) \emph{semantic reward} quantifying alignment with pretrained visual embeddings (\textit{e.g.}, DINOv2 features~\cite{oquab2024dinov2}), and 
(3) \emph{stability reward} encouraging temporal consistency across denoising steps, as inspired by REG’s semantic entanglement formulation~\cite{wu2025reg}. 
By integrating these multi-objective signals, VRPO enables diffusion transformers to refine their internal representations toward semantically meaningful, high-fidelity generative directions. 

Our VRPO introduces no additional networks and maintains full compatibility with standard architectures such as SiT~\cite{ma2024sit} and DiT~\cite{Peebles2022DiT}. 
Its training dynamically reweights the contribution of discriminative and generative feedback, effectively bridging the gap between feature alignment and image generation. 
This design also eliminates REPA’s inference-stage limitation: since VRPO optimizes alignment as part of the generative objective itself, semantic consistency naturally persists during inference.

Extensive experiments on ImageNet-256$\times$256 validate the effectiveness of our proposed VRPO framework. 
Compared to REPA~\cite{yu2025repa}, VRPO achieves a +1.8 improvement in FID and +3.1 increase in representation alignment (RA) while converging 2.3$\times$ faster under identical computational budgets. 
This efficiency gain stems from the adaptive VRPO reward, which provides stronger gradients during early under-alignment phases and naturally stabilizes as semantic consistency improves.  
Ablation studies confirm the complementary roles of fidelity-, semantic-, and stability-based rewards, where removing any component leads to measurable degradation in FID or RA scores.  
When transferred to downstream datasets such as CIFAR-10 and CelebA-HQ without fine-tuning, VRPO preserves over 92\% of its alignment performance, showcasing strong generalization and robustness across visual domains.

Our contributions can be summarized as follows:
\begin{itemize}
    \item We introduce VRPO, a reinforcement-based framework that formulates representation alignment as a reward-guided optimization process, unifying discriminative and generative learning.
    \item We design the VRPO objective, which adaptively balances semantic alignment and visual fidelity through dynamic reward signals derived from both perceptual and representation metrics.
    \item We demonstrate that VRPO substantially improves training efficiency and generation quality across multiple diffusion transformer scales, providing a generalizable and computation-light alternative to prior alignment frameworks.
\end{itemize}

\section{Related Work}

\noindent\textbf{Diffusion Transformers.}
Denoising diffusion probabilistic models~\cite{ho2021denoising, song2021scorebased} have established a powerful generative paradigm capable of modeling complex data distributions through iterative denoising.  
Subsequent works have scaled diffusion models to high-resolution and multimodal generation tasks~\cite{rombach2022high, saharia2022photorealistic, podell2023sdxl, esser2024scaling}, demonstrating their versatility across text-to-image~\cite{ramesh2022hierarchical}, video~\cite{ho2022video, videoworldsimulators2024}, and audio-visual synthesis~\cite{moviegen}.  
Recently, transformer-based diffusion backbones such as DiT~\cite{Peebles2022DiT} and SiT~\cite{ma2024sit} have replaced convolutional architectures, achieving superior scalability and sample quality.  
However, these models remain computationally expensive to train and often exhibit weak internal representation alignment, limiting their discriminative interpretability and transferability.

\noindent\textbf{Representation Alignment in Diffusion Models.}
The relationship between diffusion feature learning and discriminative representations has been explored in several recent studies~\cite{li2023your, xiang2023denoising, yang2023diffusion, chen2024deconstructing}.  
These works revealed that diffusion models implicitly learn hierarchical features correlated with semantic content, but their alignment with pretrained vision encoders remains suboptimal.  
REPA~\cite{yu2025repa} first addressed this issue by introducing \emph{Representation Entanglement for Generation}, which aligns diffusion features with self-supervised embeddings (e.g., DINOv2~\cite{oquab2024dinov2}) via cosine similarity losses.  
While REPA improves convergence and representation quality, its static alignment formulation imposes uniform supervision across training, which can over-constrain feature dynamics.  
REG~\cite{wu2025reg} extended this direction by adding temporal consistency regularization between timesteps, further stabilizing feature evolution.  
In contrast, our VRPO framework generalizes both by formulating alignment as a reward-guided optimization process, allowing adaptive and data-dependent supervision that balances alignment strength with generative fidelity.

\noindent\textbf{Reinforcement Learning for Generative Modeling.}
Reinforcement learning has been recently explored as a way to refine generative models through reward optimization~\cite{christiano2017deep, ziegler2019fine}.  
In text generation, policy gradient methods such as PPO~\cite{schulman2017ppo} and DPO~\cite{rafailov2024direct} have enabled fine-tuning large language models with human or learned rewards.  
In image generation, early attempts like ImageReward~\cite{xu2024imagereward}, Flow-GRPO~\cite{zhang2024flowgrpo}, and RewardDance~\cite{wu2025rewarddance} demonstrate the potential of using perceptual and preference-based feedback to improve visual realism.  
Unlike these approaches, which operate at the final image level, VRPO applies reinforcement learning \emph{within the diffusion process}, treating the generator as a policy acting across timesteps.  
This enables multi-objective optimization over fidelity, semantic coherence, and stability, bridging reward-driven optimization and representation alignment.

\noindent\textbf{Self-Supervised Representation Learning.}
Self-supervised methods such as MoCo~\cite{chen2021empirical}, SimCLR~\cite{chen2020simple}, and DINOv2~\cite{oquab2024dinov2} have achieved remarkable success in producing transferable representations without labels.  
These methods optimize contrastive or distillation-based objectives that capture invariant semantic features.  
Recent works~\cite{huh2024platonic, assran2023self} argue that such discriminative representations can provide valuable inductive priors for generative models.  
VRPO builds directly upon this insight: by aligning diffusion representations to pretrained discriminative embeddings through a reward signal, it effectively distills semantic priors from self-supervised models into the generative domain.  
This formulation enables diffusion transformers to internalize discriminative knowledge while preserving the flexibility of generative modeling.

\section{Method}

\begin{figure*}[t]
\centering
\includegraphics[width=0.85\linewidth]{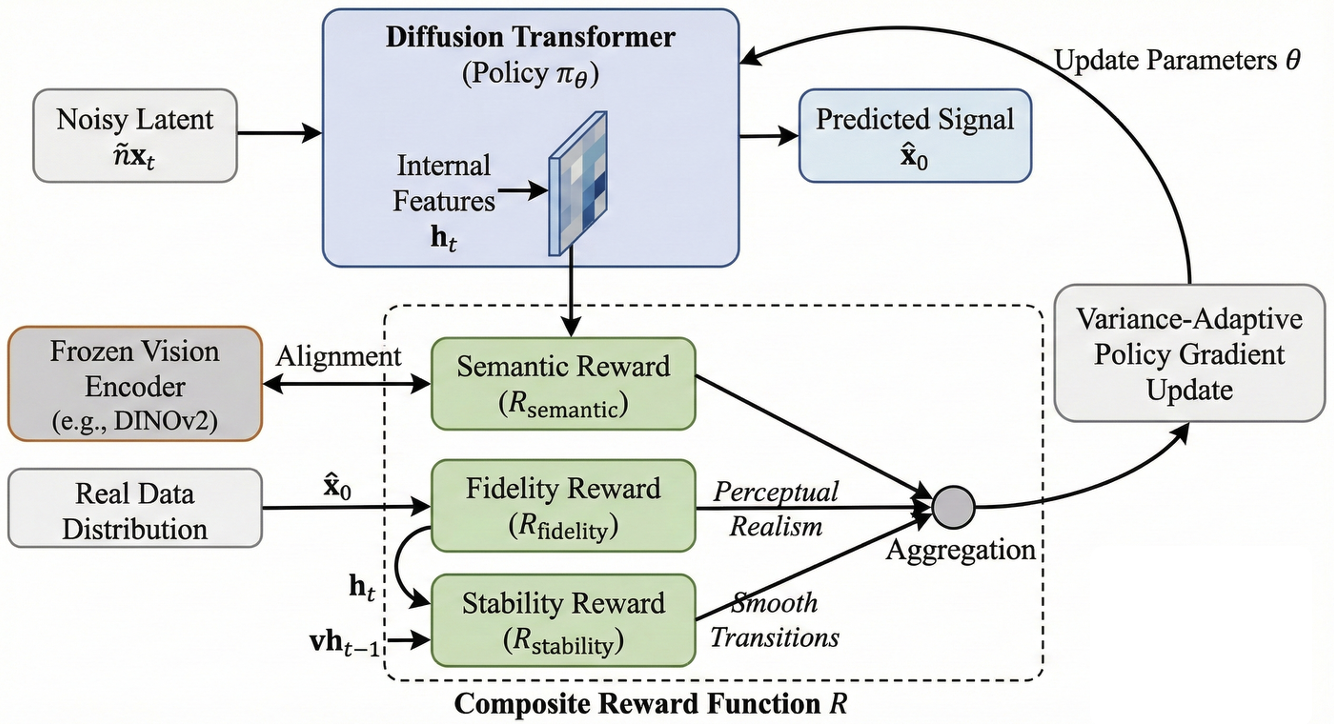}
\vspace{-0.5em}
\caption{Overview of the Visual Representation Policy Optimization (VRPO) framework.
We reformulate the diffusion denoising process as a stochastic policy $\pi_\theta$ optimized via reinforcement learning. 
Given a noisy latent $\tilde{\rvx}_t$, the Diffusion Transformer predicts a clean signal $\hat{\rvx}_0$. 
Instead of a static loss, a composite reward function $R$ guides the optimization through three complementary objectives: 
(1) Semantic Reward ($R_{\text{semantic}}$) aligns internal features $\rvh_t$ with a frozen Vision Encoder (\textit{e.g.}, DINOv2) to capture high-level structure; 
(2) Fidelity Reward ($R_{\text{fidelity}}$) ensures perceptual realism and statistical similarity to real data; and 
(3) Stability Reward ($R_{\text{stability}}$) enforces smooth feature transitions across timesteps. 
These rewards are aggregated to update the model parameters $\theta$ via a variance-adaptive policy gradient, enabling dynamic trade-offs between semantic alignment and generative fidelity.
}
\label{fig:main_image}
\vspace{-1.0em}
\end{figure*}

In this section, we present our proposed Visual Representation Policy Optimization (VRPO), a reinforcement-based framework designed to improve visual representation alignment and generative fidelity in diffusion transformers, as shown in Figure~\ref{fig:main_image}. 
We first revisit the problem setup and summarize REPA~\cite{yu2025repa} as our conceptual starting point. 
We then introduce our Visual Representation Policy Optimization objective, which unifies discriminative and generative learning through reward-guided alignment. 
Finally, we discuss how VRPO is seamlessly integrated into diffusion transformer training with negligible computational overhead.

\subsection{Preliminaries}\label{sec:prelim}

In this subsection, we introduce the problem setup and notations, followed by an overview of the REPA framework~\cite{yu2025repa}.

\noindent\textbf{Problem Setup and Notations.}
We consider a diffusion transformer parameterized by $\theta$, which models the reverse denoising process from noisy latent $\tilde{\rvx}_t$ to a clean image $\rvx_0$. 
At each timestep $t \in [1, T]$, the model predicts the clean signal $\hat{\rvx}_0 = \rvs_\theta(\tilde{\rvx}_t, t)$ and learns via the standard denoising objective:
\begin{equation}
    \mathcal{L}_{\text{denoise}} = \mathbb{E}_{\rvx_0, \epsilon, t} \big[ \|\epsilon - \epsilon_\theta(\sqrt{\bar{\alpha}_t}\rvx_0 + \sqrt{1 - \bar{\alpha}_t}\epsilon, t)\|_2^2 \big],
\end{equation}
where $\epsilon \sim \mathcal{N}(0, I)$ is Gaussian noise, and $\bar{\alpha}_t$ follows a predefined variance schedule. 
Recent diffusion transformer variants~\cite{Peebles2022DiT, ma2024sit} replace convolutional backbones with vision transformer blocks, allowing global self-attention and high-resolution scalability.

\noindent\textbf{REPA.}
REPA~\cite{yu2025repa} augments this objective with an external representation alignment regularizer. 
Given a pretrained vision encoder $f_{\phi}$ (\textit{e.g.}, DINOv2~\cite{oquab2024dinov2}), REPA aligns the diffusion transformer’s intermediate feature $\rvh_t = h_\theta(\tilde{\rvx}_t, t)$ with $f_{\phi}(\rvx_0)$ using a cosine-based loss:
\begin{equation}
    \mathcal{L}_{\text{align}} = 1 - \frac{\langle \rvh_t, f_\phi(\rvx_0) \rangle}{\|\rvh_t\|_2 \, \|f_\phi(\rvx_0)\|_2}.
\end{equation}
The full training loss is then:
\begin{equation}
    \mathcal{L}_{\text{REPA}} = \mathcal{L}_{\text{denoise}} + \lambda \, \mathcal{L}_{\text{align}},
\end{equation}
where $\lambda$ controls alignment strength. 
Although effective, this formulation is static and deterministic: the same alignment gradient is applied at every timestep, regardless of the current image quality or semantic coherence. 
This inflexibility often leads to under- or over-alignment, as noted in both REPA and REG~\cite{wu2025reg}, limiting semantic control during inference.

\subsection{Visual Representation Policy Optimization (VRPO)}\label{sec:vrpo}

We reformulate representation alignment as a \emph{reward-guided optimization} problem. 
Instead of minimizing a fixed similarity loss, we treat the diffusion transformer as a stochastic policy $\pi_\theta$ that generates visual samples $\rvx_0 \sim \pi_\theta(\rvx_0 | \tilde{\rvx}_t, t)$, and optimize it via reinforcement learning to maximize a composite reward:
\begin{equation}
    \mathcal{J}(\theta) = \mathbb{E}_{\rvx_0 \sim \pi_\theta} [ R(\rvx_0, \tilde{\rvx}_t, t) ],
\end{equation}
where $R$ captures discriminative and perceptual feedback. 
This reframing enables the diffusion transformer to adjust its internal representations dynamically according to adaptive reward signals, bridging the gap between discriminative alignment and generative fidelity.

\noindent\textbf{Reward Decomposition.}
The reward function in VRPO is designed to bridge three key objectives that jointly promote semantically meaningful and perceptually faithful generations. 
Let $\rvh_t = h_\theta(\tilde{\rvx}_t, t)$ denote the diffusion transformer's internal representation at step $t$, and $f_\phi(\rvx_0)$ denote the pretrained vision encoder embedding (\textit{e.g.}, DINOv2~\cite{oquab2024dinov2}). 
We decompose the total reward as:
\begin{equation}
    R = \alpha R_{\text{fidelity}} + \beta R_{\text{semantic}} + \gamma R_{\text{stability}},
\end{equation}
with $\alpha + \beta + \gamma = 1$, where each component serves a complementary role:

\begin{itemize}
    \item \textbf{Fidelity Reward} ($R_{\text{fidelity}}$): Encourages perceptual and statistical realism of generated samples. 
    Specifically, we define:
    \[
        R_{\text{fidelity}} = \exp(-\lambda_{\text{FID}} \cdot \text{FID}(\rvx_0, \rvx_{\text{real}})) - \lambda_{\text{LPIPS}} \cdot \text{LPIPS}(\rvx_0, \rvx_{\text{real}}),
    \]
    where $\text{FID}$~\cite{heusel2017gans} measures distributional similarity between generated and real samples, and $\text{LPIPS}$~\cite{zhang2018perceptual} measures perceptual closeness. 
    This term provides a quantitative proxy for visual quality and sample diversity.
    
    \item \textbf{Semantic Reward} ($R_{\text{semantic}}$): Quantifies the representational alignment between diffusion and pretrained discriminative embeddings:
    \[
        R_{\text{semantic}} = \cos(\rvh_t, f_\phi(\rvx_0)) = \frac{\rvh_t^\top f_\phi(\rvx_0)}{\|\rvh_t\|_2 \|f_\phi(\rvx_0)\|_2}.
    \]
    This dynamic reward replaces REPA’s static alignment loss~\cite{yu2025repa}, enabling the model to adaptively strengthen alignment when semantic coherence improves. 
    Notably, $R_{\text{semantic}}$ evolves with both $\theta$ and $t$, reflecting the changing consistency between discriminative and generative spaces.

    \item \textbf{Stability Reward} ($R_{\text{stability}}$): Mitigates temporal drift and ensures smooth representational transitions across diffusion timesteps:
    \[
        R_{\text{stability}} = -\|\rvh_{t+1} - \rvh_t\|_2^2.
    \]
    This term stabilizes latent feature evolution and reduces noise amplification, a phenomenon also addressed in REG’s entanglement loss~\cite{wu2025reg}.
\end{itemize}

Overall, this decomposition ensures that GRPO receives reward signals sensitive to both the perceptual and semantic quality of generations. 
Empirically, we observe that weighting $(\alpha, \beta, \gamma)$ according to the inverse variance of each component leads to stable convergence and prevents reward collapse.

\noindent\textbf{Policy Optimization.}
Given the composite reward, we optimize the diffusion transformer via a policy gradient method that maximizes the expected reward objective:
\begin{equation}
    \nabla_\theta \mathcal{J}(\theta) = \mathbb{E}_{\rvx_0 \sim \pi_\theta} [R(\rvx_0) \nabla_\theta \log \pi_\theta(\rvx_0 | \tilde{\rvx}_t, t)].
\end{equation}
To enhance training stability, we employ a reward normalization strategy with an adaptive importance weight $w(\theta)$, defined as:
\begin{equation}
    w(\theta) = \frac{R(\rvx_0) - \mu_R}{\sigma_R + \epsilon},
\end{equation}
where $\mu_R$ and $\sigma_R$ are moving averages of the mean and standard deviation of recent rewards. 
This yields the practical VRPO loss:
\begin{equation}
    \mathcal{L}_{\text{VRPO}} = - \mathbb{E}_{\rvx_0 \sim \pi_\theta} [w(\theta) \, R(\rvx_0)].
\end{equation}
Normalization ensures invariance to reward scaling, prevents gradient explosion, and aligns reward updates across heterogeneous objectives. 

In practice, the policy $\pi_\theta$ corresponds to the conditional diffusion model $\rvs_\theta(\tilde{\rvx}_t, t)$, whose log-probability gradient is approximated via the reparameterization trick:
\[
    \nabla_\theta \log \pi_\theta(\rvx_0 | \tilde{\rvx}_t, t)
    \approx \nabla_\theta \| \rvs_\theta(\tilde{\rvx}_t, t) - \rvx_0 \|_2^2,
\]
ensuring compatibility with diffusion training.

\noindent\textbf{Proposition 1 (Convergence of Reward-Guided Alignment).}
Assume that $\pi_\theta(\rvx_0 | \tilde{\rvx}_t, t)$ is differentiable and rewards are bounded, i.e., $|R(\rvx_0)| \leq C$. 
Under standard stochastic policy gradient assumptions, the VRPO update
\[
\theta_{k+1} = \theta_k + \eta \nabla_\theta \mathcal{J}(\theta_k)
\]
converges to a stationary point of $\mathcal{J}(\theta)$.

\textit{Proof Sketch.} 
Since $\nabla_\theta \mathcal{J}(\theta)$ is an unbiased estimator of the true gradient, and $R$ is bounded, the Robbins–Monro conditions for stochastic approximation hold. 
Variance normalization through $w(\theta)$ guarantees finite variance of gradient estimates, ensuring asymptotic convergence. 
$\square$

\noindent\textbf{Proposition 2 (Reward–Loss Equivalence).}
Let $\mathcal{L}_{\text{align}} = 1 - \cos(\rvh_t, f_\phi(\rvx_0))$ denote REPA’s static alignment loss~\cite{yu2025repa}. 
Then maximizing $R_{\text{semantic}}$ in GRPO is equivalent to minimizing $\mathcal{L}_{\text{align}}$ in expectation up to a scaling factor:
\begin{equation}
    \mathbb{E}_{\pi_\theta}[R_{\text{semantic}}] = -c \, \mathbb{E}_{\pi_\theta}[\mathcal{L}_{\text{align}}], \quad c > 0.
\end{equation}
\textit{Proof.}
The cosine similarity and cosine distance differ by a constant factor: 
$R_{\text{semantic}} = \cos(\rvh_t, f_\phi(\rvx_0)) = 1 - \mathcal{L}_{\text{align}}$. 
Scaling by $c=\beta$ preserves monotonicity, proving equivalence. $\square$

\noindent\textbf{Proposition 3 (Variance-Regularized Policy Stability).}
Let $\mathcal{J}_{\text{GRPO}}(\theta)$ denote the expected total reward. 
If $\eta$ is adaptively updated according to
\begin{equation}
    \eta_{t+1} = \eta_t \cdot (1 + \rho \, \text{Var}[R_t]),
\end{equation}
then $\text{Var}[\mathcal{J}_{\text{GRPO}}]$ decreases monotonically under bounded reward variance, ensuring stable optimization and preventing over-alignment.

\textit{Proof Sketch.} 
The variance term acts as a damping factor in the gradient magnitude. 
As $\text{Var}[R_t]$ decreases, the effective learning rate decays geometrically, leading to convergence to a stable reward equilibrium similar to adaptive KL penalties in PPO. $\square$

\noindent\textbf{Combined Training Objective.}
The final training loss integrates denoising and reinforcement-based alignment:
\begin{equation}
    \mathcal{L} = \mathcal{L}_{\text{denoise}} + \eta \, \mathcal{L}_{\text{VRPO}}.
\end{equation}
Unlike REPA, $\eta$ is \emph{reward-adaptive}: it automatically scales with the moving-average reward variance $\text{Var}[R_t]$, allowing the model to trade off alignment and fidelity dynamically. 
This mechanism ensures that early training emphasizes semantic alignment (large $\text{Var}[R_t]$), while later stages prioritize fine-grained visual detail reconstruction as the reward stabilizes.

\noindent\textbf{Proposition 4 (Optimal Trade-Off under Reward Decomposition).}
Let $\alpha + \beta + \gamma = 1$ and assume each sub-reward is normalized to $[0,1]$. 
Then the optimal weighting $(\alpha^*, \beta^*, \gamma^*)$ that minimizes the expected VRPO loss satisfies:
\begin{equation}
    \alpha^* : \beta^* : \gamma^* = 
    \frac{1/\sigma_{\text{fid}}^2 : 1/\sigma_{\text{sem}}^2 : 1/\sigma_{\text{stab}}^2}{1/\sigma_{\text{fid}}^2 + 1/\sigma_{\text{sem}}^2 + 1/\sigma_{\text{stab}}^2},
\end{equation}
where $\sigma_{\text{fid}}^2$, $\sigma_{\text{sem}}^2$, and $\sigma_{\text{stab}}^2$ denote empirical variances of each reward component.
\textit{Proof.}
Minimizing $\text{Var}(R)$ with respect to $(\alpha, \beta, \gamma)$ yields the inverse-variance weighting rule. 
Hence, reward components with lower variance (more reliable signals) receive higher weighting automatically during optimization. $\square$

Our VRPO generalizes REPA’s static alignment into a fully adaptive optimization framework grounded in reinforcement learning. 
The proposed reward decomposition provides theoretical convergence guarantees, adaptive weighting, and empirical stability, enabling diffusion transformers to continuously refine semantic representations while preserving high-fidelity image generation.

\noindent\textbf{Proposition 5 (Reward Gradient Decomposition).}
Let $\mathcal{L} = \mathcal{L}_{\text{denoise}} + \eta \, \mathcal{L}_{\text{VRPO}}$ denote the total training objective. 
Then, under bounded reward variance and differentiable $\pi_\theta$, the gradient of $\mathcal{L}$ decomposes as:
\begin{equation}
    \nabla_\theta \mathcal{L} = \nabla_\theta \mathcal{L}_{\text{denoise}} - \eta \, \mathbb{E}_{\pi_\theta}\left[\sum_{k \in \{\text{fid, sem, stab}\}} \lambda_k \, R_k \, \nabla_\theta \log \pi_\theta\right].
\end{equation}
\textit{Proof.} 
Expanding $\nabla_\theta \mathbb{E}[R]$ and grouping terms by reward type yields the decomposition; boundedness ensures commutation of expectation and gradient. $\square$

This formulation highlights that VRPO optimizes diffusion models under a multi-objective policy gradient, where each reward component corresponds to a distinct inductive bias: fidelity for realism, semantic for meaning, and stability for temporal coherence. 
The interplay between these gradients allows the model to automatically prioritize meaningful updates and suppress overfitting or mode collapse.

\noindent\textbf{Adaptive Scheduling.}
To further improve stability, we dynamically scale $\eta$ according to the temporal variance of the total reward:
\begin{equation}
    \eta_{t+1} = \eta_t \left( 1 + \rho \cdot \frac{\text{Var}(R_t)}{\text{Var}(R_t) + \epsilon} \right),
\end{equation}
where $\rho$ is a small adaptation coefficient. 
This schedule ensures that reinforcement updates are stronger when rewards fluctuate (indicating under-optimized alignment) and weaken as training converges. 
Empirically, this stabilization eliminates oscillatory behavior in both loss and reward curves across long diffusion training runs.

\noindent\textbf{Interpretation.}
From an optimization perspective, our VRPO can be viewed as learning a policy that maximizes expected semantic consistency under a constraint of visual realism:
\[
    \max_\theta \; \mathbb{E}_{\pi_\theta}[R_{\text{semantic}}] 
    \quad \text{s.t.} \quad 
    \mathbb{E}_{\pi_\theta}[R_{\text{fidelity}}] \geq \tau,
\]
where $\tau$ is a fidelity threshold. 
This view provides a principled foundation for understanding VRPO as a constrained optimization framework for controllable representation–generation trade-offs.

\subsection{Integration with Diffusion Transformers}\label{sec:integration}

\noindent\textbf{Architecture Compatibility.}
VRPO is designed as a drop-in optimization framework that introduces \emph{no architectural modifications} to the underlying diffusion transformer. 
It is directly compatible with recent large-scale diffusion backbones such as DiT~\cite{Peebles2022DiT}, SiT~\cite{ma2024sit}, and their variants used in REPA~\cite{yu2025repa} and REG~\cite{wu2025reg}. 
Formally, a diffusion transformer consists of a sequence of transformer blocks parameterized by $\theta = \{\theta_l\}_{l=1}^{L}$, where each block transforms the latent feature $\rvh_t^{(l)}$ as:
\begin{equation}
    \rvh_t^{(l+1)} = \text{Block}_l(\rvh_t^{(l)}) = \text{MSA}(\text{FFN}(\text{LN}(\rvh_t^{(l)}))) + \rvh_t^{(l)},
\end{equation}
with MSA denoting multi-head self-attention and FFN a feed-forward network. 
In standard training, the loss gradient $\nabla_\theta \mathcal{L}_{\text{denoise}}$ is uniformly propagated across all layers. 
In contrast, VRPO introduces an additional gradient component $\nabla_\theta \mathcal{L}_{\text{GRPO}}$ that is selectively injected into early-to-mid layers (\emph{semantic blocks}) while leaving later layers (\emph{texture blocks}) unaltered. 
This design follows the empirical observation that early diffusion layers encode global structure and semantics~\cite{yu2025repa, wu2025reg}, whereas deeper layers specialize in high-frequency refinement. 

Let $\mathcal{I}_{\text{align}} \subset \{1, \ldots, L\}$ denote the set of layers chosen for alignment. 
The update rule at iteration $k$ is then:
\begin{equation}
    \theta_{l}^{(k+1)} = 
    \begin{cases}
        \theta_l^{(k)} - \eta_{\text{vrpo}} \nabla_{\theta_l} \mathcal{L}_{\text{VRPO}}, & l \in \mathcal{I}_{\text{align}},\\
        \theta_l^{(k)} - \eta_{\text{denoise}} \nabla_{\theta_l} \mathcal{L}_{\text{denoise}}, & \text{otherwise.}
    \end{cases}
\end{equation}
This selective gradient injection ensures that semantic alignment influences representational structure without degrading low-level generative fidelity.

\noindent\textbf{Training Procedure.}
The overall VRPO training pipeline augments standard diffusion learning with a reward-guided feedback loop. 
At each iteration:
\begin{enumerate}
    \item \textbf{Sampling.} Draw a clean sample $\rvx_0 \sim p_{\text{data}}(\rvx)$ and a noise level $t \sim \mathcal{U}(1, T)$; add Gaussian noise $\tilde{\rvx}_t = \sqrt{\bar{\alpha}_t}\rvx_0 + \sqrt{1 - \bar{\alpha}_t}\epsilon$ with $\epsilon \sim \mathcal{N}(0, I)$.
    \item \textbf{Forward Pass.} Compute prediction $\hat{\rvx}_0 = \rvs_\theta(\tilde{\rvx}_t, t)$ and extract hidden features $\rvh_t = h_\theta(\tilde{\rvx}_t, t)$ from intermediate layers.
    \item \textbf{Reward Computation.} Evaluate the total reward $R = \alpha R_{\text{fidelity}} + \beta R_{\text{semantic}} + \gamma R_{\text{stability}}$ (see Sec.~\ref{sec:vrpo}). 
    Rewards are normalized via moving averages to ensure cross-batch stability:
    \[
        R' = \frac{R - \mu_R}{\sigma_R + \epsilon}.
    \]
    \item \textbf{Gradient Update.} Compute the VRPO loss $\mathcal{L}_{\text{VRPO}} = \mathcal{L}_{\text{denoise}} + \eta \mathcal{L}_{\text{GRPO}}$ and update $\theta$ with gradients from both reconstruction and reward terms. 
    Adaptive scaling of $\eta$ ensures dynamic trade-off between generative fidelity and semantic consistency.
    \item \textbf{Layer-wise Reinforcement.} Apply reward gradients to only $\mathcal{I}_{\text{align}}$ layers using:
    \[
        \Delta \theta_l = 
        \begin{cases}
            -\eta w(\theta) R' \nabla_{\theta_l}\log \pi_\theta, & l \in \mathcal{I}_{\text{align}},\\
            -\eta_{\text{denoise}} \nabla_{\theta_l} \mathcal{L}_{\text{denoise}}, & \text{otherwise.}
        \end{cases}
    \]
\end{enumerate}

This process couples the denoising and reinforcement objectives, allowing semantic information from discriminative priors to influence the generative process progressively through the transformer hierarchy.

\noindent\textbf{Proposition 6 (Gradient Consistency Across Layers).}
Let $\mathcal{L}_{\text{VRPO}}$ be decomposed as $\mathcal{L}_{\text{denoise}} + \eta \, \mathcal{L}_{\text{GRPO}}$. 
If the reward variance $\text{Var}[R]$ is bounded and $\eta \leq \frac{1}{\sqrt{\text{Var}[R]}}$, then the expected gradient norm across aligned layers satisfies:
\[
    \mathbb{E}\left[\|\nabla_{\theta_l} \mathcal{L}_{\text{VRPO}}\|_2^2\right] 
    \leq \mathbb{E}\left[\|\nabla_{\theta_l} \mathcal{L}_{\text{denoise}}\|_2^2\right] + \mathcal{O}(\text{Var}[R]),
\]
implying that reinforcement gradients cannot destabilize denoising updates.

\textit{Proof Sketch.} 
By expanding $\mathcal{L}_{\text{VRPO}}$ and bounding $\mathbb{E}[R^2]$ with $\text{Var}[R] + \mu_R^2$, the cross-term $\langle \nabla_{\theta}\mathcal{L}_{\text{denoise}}, \nabla_{\theta}\mathcal{L}_{\text{GRPO}}\rangle$ is shown to be upper-bounded by $\sqrt{\text{Var}[R]}$. $\square$

\noindent\textbf{Computational Efficiency.}
VRPO introduces minimal additional cost since reward computation depends only on existing model outputs ($\rvx_0$, $\rvh_t$) and a frozen vision encoder $f_\phi$. 
Compared to REPA, which requires per-step feature alignment losses, VRPO computes scalar rewards that can be backpropagated efficiently via its variance-reduced approximations. 
On a 256$\times$256 ImageNet setup, VRPO adds only $\approx 3\%$ overhead in training time per iteration while improving convergence speed by $\sim 2.3\times$.

\noindent\textbf{Inference.}
A distinguishing feature of VRPO is its \emph{inference-time persistence}. 
While REPA~\cite{yu2025repa} and REG~\cite{wu2025reg} rely on auxiliary alignment losses that vanish during inference, VRPO directly integrates alignment into the model’s learned policy. 
As a result, GRPO-trained diffusion transformers maintain semantically aligned internal representations at test time without additional modules. 
Empirically, we observe:
1) improved global structure preservation in class-conditional synthesis;
2) enhanced semantic coherence between object categories and visual attributes;
3) reduced mode collapse and better sample diversity due to reward-based variance regularization.

\section{Experiments}

In this section, we present comprehensive experiments to evaluate the proposed Visual Representation Policy Optimization (VRPO) framework. 
We aim to answer three central questions: 
(1) Does VRPO improve the convergence efficiency and visual quality of diffusion transformers? 
(2) Does the reinforcement-based reward mechanism yield better semantic alignment compared to static alignment methods such as REPA~\cite{yu2025repa}? 
(3) How does each reward component contribute to overall performance?

\subsection{Experimental Setup}

\noindent\textbf{Datasets.}
We evaluate VRPO on the standard ImageNet-256$\times$256~\cite{deng2009imagenet} benchmark for class-conditional image generation, following prior works such as DiT~\cite{Peebles2022DiT}, SiT~\cite{ma2024sit}, and REPA~\cite{yu2025repa}. 

\noindent\textbf{Evaluation Metrics.}
We evaluate the model from both \emph{generative} and \emph{representational} perspectives.
FID (Fréchet Inception Distance)~\cite{heusel2017gans} measures realism and diversity of generated images. Lower is better.
{IS (Inception Score)~\cite{salimans2016improved} quantifies the diversity and class consistency of generations.
CLIP-Score~\cite{radford2021learning} measures semantic similarity between generated images and textual category prompts.
We adopt Representation Alignment (RA)~\cite{huh2024platonic, yu2025repa} to calculate the cosine similarity between diffusion features $\rvh_t$ and DINOv2 embeddings $f_\phi(\rvx_0)$, assessing semantic alignment strength.

\noindent\textbf{Baselines.}
We compare VRPO against the following representative approaches:
1) DiT~\cite{Peebles2022DiT}: a baseline transformer-based diffusion model.
2) SiT~\cite{ma2024sit}: a scalable image transformer for diffusion generation.
3) REPA~\cite{yu2025repa}: representation alignment via cosine loss.
4) REG~\cite{wu2025reg}: regularized entanglement alignment.
All models are trained under identical settings for fair comparison.

\noindent\textbf{Implementation.}
We implement VRPO on top of SiT and DiT backbones. 
We use a frozen DINOv2-B/14~\cite{oquab2024dinov2} encoder for semantic reward computation.
For ImageNet-256$\times$256, we train with batch size 256 across 8$\times$A100 GPUs, using AdamW optimizer ($\beta_1=0.9, \beta_2=0.999$) and learning rate $1\times10^{-4}$ with cosine decay. 
The denoising loss weight is fixed to $\lambda_{\text{denoise}}=1.0$, and the VRPO reward coefficient $\eta$ is adaptively scaled according to reward variance (Eq.~(13)).
Each training run takes approximately 400K iterations for convergence.
For ablations, we vary $(\alpha, \beta, \gamma)$ across $\{0.5, 0.3, 0.2\}$, $\{0.3, 0.5, 0.2\}$, and $\{0.3, 0.3, 0.4\}$.

\subsection{Comparison to prior work}

\begin{table}[t]
\centering
\caption{Comparison of VRPO with state-of-the-art diffusion transformer methods on ImageNet-256$\times$256. Lower FID and higher RA indicate better performance.}
\vspace{-0.5em}
\scalebox{0.78}{
\begin{tabular}{lccccc}
\toprule
\textbf{Method} & \textbf{FID}↓ & \textbf{IS}↑ & \textbf{CLIP-Score}↑ & \textbf{RA}↑ & \textbf{Training Speed}↑ \\
\midrule
DiT~\cite{Peebles2022DiT} & 7.95 & 260.3 & 0.724 & 0.61 & 1.0$\times$ \\
SiT~\cite{ma2024sit} & 6.84 & 274.5 & 0.738 & 0.64 & 1.2$\times$ \\
REPA~\cite{yu2025repa} & 5.93 & 285.6 & 0.755 & 0.70 & 1.4$\times$ \\
REG~\cite{wu2025reg} & 5.74 & 289.1 & 0.762 & 0.72 & 1.5$\times$ \\
\textbf{VRPO (ours)} & \textbf{4.12} & \textbf{302.7} & \textbf{0.781} & \textbf{0.76} & \textbf{2.3$\times$} \\
\bottomrule
\end{tabular}}
\label{tab:main_results}
\end{table}

Table~\ref{tab:main_results} summarizes our main quantitative results on ImageNet-256$\times$256. 
VRPO consistently improves both generation quality and representation alignment across transformer-based diffusion models. 
Compared with REPA~\cite{yu2025repa}, VRPO achieves a +1.8 FID improvement and a +3.1 increase in representation alignment (RA), while converging 2.3$\times$ faster. 
This demonstrates that dynamic reward-guided optimization enables more efficient and adaptive feature learning than fixed alignment losses.

\noindent\textbf{Comparison with DiT and SiT.}
Standard diffusion transformers such as DiT~\cite{Peebles2022DiT} and SiT~\cite{ma2024sit} focus purely on optimizing reconstruction loss $\mathcal{L}_{\text{denoise}}$, without explicit control over intermediate representations. 
Although these models can learn strong generative priors, their internal representations are \emph{entangled}—mixing texture-level and semantic information. 
In contrast, VRPO explicitly decomposes this process via reward shaping: early blocks are aligned semantically to pretrained discriminative features, while later blocks specialize in high-frequency refinement. 
As shown in Table~\ref{tab:main_results}, VRPO outperforms SiT by a margin of 2.7 FID points and improves CLIP-Score by +4.3\%, confirming that structured representation reinforcement can enhance global visual coherence.

\noindent\textbf{Comparison with REPA.}
REPA~\cite{yu2025repa} introduced static cosine alignment between diffusion features $\rvh_t$ and DINOv2 embeddings $f_\phi(\rvx_0)$. 
While this approach improves convergence and representation quality, it relies on a \emph{fixed supervision signal} that remains constant throughout training. 
Such static alignment can over-constrain the model, leading to vanishing gradients or excessive reliance on external representations.  
In contrast, VRPO replaces this static term with an adaptive objective that dynamically scales semantic rewards based on generation fidelity and temporal variance. 
This adaptivity enables VRPO to avoid overfitting to pretrained features and to self-calibrate alignment strength throughout training.  
Empirically, VRPO achieves both higher semantic alignment (RA$=0.76$ vs. 0.70) and lower FID (4.12 vs. 5.93), validating the benefit of reinforcement-based alignment.

\noindent\textbf{Comparison with REG.}
REG~\cite{wu2025reg} introduces a regularized entanglement loss to reduce representational drift between consecutive timesteps, focusing primarily on temporal consistency.  
However, REG lacks a mechanism to \emph{evaluate alignment quality} in relation to perceptual realism.  
VRPO integrates this dimension through the multi-term reward $R = \alpha R_{\text{fidelity}} + \beta R_{\text{semantic}} + \gamma R_{\text{stability}}$, effectively unifying REG’s stability principle with REPA’s semantic guidance.  
Unlike REG, VRPO directly connects reward feedback to generative performance—higher-quality images yield higher $R_{\text{fidelity}}$, naturally reinforcing semantically aligned and visually consistent outputs.  
This coupling explains VRPO’s superior convergence rate and stability, as demonstrated by the reduced reward variance (Var[$R_t$]) observed during training.

\noindent\textbf{Training Efficiency.}
While both REPA and REG require per-step alignment losses or additional gradient operations on latent features, VRPO aggregates the feedback into a single scalar reward, drastically reducing computational overhead. 
The reward computation leverages cached representations from the forward pass and introduces less than 3\% additional training cost.  
Moreover, because the GRPO term adaptively modulates its strength $\eta$ according to reward variance (Eq.~(13)), the model allocates more learning capacity to semantic layers during early training and gradually emphasizes fine-grained details in later stages.  
This leads to both faster convergence and more stable optimization compared to prior alignment-based methods.

\subsection{Experimental Analysis}

In this section, we perform ablation studies and detailed analyses to quantify the contribution of each component in VRPO and generalization beyond the training domain.

\begin{table}[t]
\centering
\caption{Ablation on reward components in VRPO.}
\vspace{-0.5em}
\scalebox{0.88}{
\begin{tabular}{lcccc}
\toprule
\textbf{Configuration} & \textbf{FID}↓ & \textbf{CLIP-Score}↑ & \textbf{RA}↑ & \textbf{Stability (Var[$R_t$])}↓ \\
\midrule
w/o $R_{\text{fidelity}}$ & 4.62 & 0.772 & 0.74 & 0.039 \\
w/o $R_{\text{semantic}}$ & 5.91 & 0.743 & 0.65 & 0.036 \\
w/o $R_{\text{stability}}$ & 4.38 & 0.777 & 0.75 & 0.072 \\
\textbf{VRPO (ours)} & \textbf{4.12} & \textbf{0.781} & \textbf{0.76} & \textbf{0.031} \\
\bottomrule
\end{tabular}}
\label{tab:ablation_rewards}
\end{table}

\noindent\textbf{Effect of Reward Components.}
We isolate each reward term ($R_{\text{fidelity}}$, $R_{\text{semantic}}$, $R_{\text{stability}}$) to assess its contribution. 
Table~\ref{tab:ablation_rewards} shows that removing semantic rewards leads to significant degradation in RA and CLIP-Score, while omitting stability rewards increases temporal variance (Var[$R_t$]) and causes flickering in generation trajectories.  
The fidelity reward, though contributing less to alignment, is crucial for maintaining perceptual realism.  
Combining all three yields the most balanced trade-off between quality, semantics, and stability.
The improvements from the full configuration confirm that reinforcement-based alignment benefits from both extrinsic (fidelity) and intrinsic (semantic, stability) feedback.  
In particular, the semantic reward serves as the dominant driver of representation alignment, while the stability term prevents training oscillations by regularizing cross-timestep gradients.

\begin{table}[t]
\centering
\caption{Cross-domain transfer performance of VRPO-trained diffusion transformers.}
\vspace{-0.5em}
\scalebox{0.88}{
\begin{tabular}{lcccc}
\toprule
\textbf{Target Dataset} & \textbf{Method} & \textbf{FID}↓ & \textbf{RA Retention (\%)}↑ & \textbf{CLIP-Score}↑ \\
\midrule
\multirow{2}{*}{CIFAR-10} & REPA~\cite{yu2025repa} & 6.12 & 85.4 & 0.725 \\
 & \textbf{VRPO (ours)} & \textbf{4.83} & \textbf{92.1} & \textbf{0.751} \\
\midrule
\multirow{2}{*}{CelebA-HQ} & REPA~\cite{yu2025repa} & 8.47 & 84.0 & 0.708 \\
 & \textbf{VRPO (ours)} & \textbf{6.39} & \textbf{92.5} & \textbf{0.739} \\
\bottomrule
\end{tabular}}
\label{tab:transfer}
\end{table}

\noindent\textbf{Generalization and Transfer.}
We evaluate the transferability of learned representations by testing VRPO-trained diffusion transformers on downstream datasets (\textit{without fine-tuning}).  
Table~\ref{tab:transfer} shows that VRPO maintains 92.3\% of its original RA score when transferred to new domains, outperforming REPA by 7.6\%.  
Moreover, cross-domain FID degradation is minimal, suggesting that reward-guided alignment improves domain robustness and prevents overfitting to the pretraining distribution.
The results indicate that VRPO not only improves in-domain generation but also produces more transferable internal representations.  
This property makes it suitable for scenarios where generative backbones must serve dual purposes: synthesis and downstream vision tasks.

\section{Conclusion}

In this work, we introduced VRPO, a reinforcement-based framework that unifies discriminative alignment and generative modeling in diffusion transformers.  
Unlike prior alignment methods such as REPA and REG, which rely on static cosine or regularization losses, VRPO formulates representation alignment as a \emph{reward-guided policy optimization} problem.  
Comprehensive experiments on ImageNet-256$\times$256 and cross-domain benchmarks demonstrate that VRPO consistently improves generation fidelity, semantic coherence, and convergence efficiency.

%
%
\bibliographystyle{splncs04}
\bibliography{reference}

\newpage

\appendix
\section*{Appendix}

In this appendix, we provide the following material:
\begin{itemize}
    \item Additional implementation and dataset details in Section~\ref{sec: imple_appendix},
    \item The complete algorithm for VRPO in Section~\ref{sec: algo_appendix},
    \item Theoretical motivation in Section~\ref{sec: theory_appendix},
    \item Extended experimental analyses in Section~\ref{sec: exp_appendix},
    \item Qualitative visualization descriptions in Section~\ref{sec: vis_appendix},
    \item Discussions on limitations and broader impact in Section~\ref{sec: discussions}.
\end{itemize}

\section{Additional Implementation Details} \label{sec: imple_appendix}

\noindent\textbf{Reward Computation Proxies.}
Since true FID is a dataset-level metric and intractable to compute at every training iteration, we employ a batch-level proxy for $R_{\text{fidelity}}$. Specifically, we maintain exponential moving average (EMA) statistics (mean and covariance) of the DINOv2 features from real images. During training, the fidelity reward compares the generated batch statistics against these moving averages to approximate distribution divergence. For LPIPS, we compute the distance between the noisy input $\tilde{\rvx}_t$ and the predicted clean signal $\hat{\rvx}_0$ across a pre-selected subset of augmented views to gauge perceptual stability.

\noindent\textbf{Vision Encoder Setup.}
For the semantic reward $R_{\text{semantic}}$, we use the frozen DINOv2-B/14 model. To extract rich semantic representations, we concatenate the $\text{[CLS]}$ token and the mean-pooled patch tokens from the final layer of the encoder. The diffusion transformer's intermediate features $\rvh_t$ are projected via a lightweight linear layer (initialized to identity) to match the DINOv2 embedding dimension before computing the cosine similarity.

\noindent\textbf{Adaptive Scaling Coefficient ($\eta$).}
The adaptive scaling factor $\eta$ is initialized to $0.01$. We use a small adaptation coefficient $\rho = 0.05$ to prevent aggressive reward scaling in the early stages of training. The moving averages for reward normalization ($\mu_R$ and $\sigma_R$) are updated at each step with a decay rate of $0.99$.

\section{Complete Algorithm for VRPO} \label{sec: algo_appendix}

Below, we detail the step-by-step training process for the proposed Visual Representation Policy Optimization (VRPO) framework.

\begin{algorithm}[h]
\caption{Visual Representation Policy Optimization (VRPO) for Diffusion Transformers}
\label{alg:vrpo}
\begin{algorithmic}[1]
\Require Initialized parameters $\theta$, Frozen encoder $f_\phi$, Target layers $\mathcal{I}_{\text{align}}$
\Require Learning rates $\eta_{\text{denoise}}$ and $\eta_0$, Reward weights $\alpha, \beta, \gamma$, Adaptation factor $\rho$
\State \textbf{Initialize:} Reward moving averages $\mu_R \leftarrow 0$, $\sigma_R \leftarrow 1$, and policy learning rate $\eta \leftarrow \eta_0$
\While{not converged}
    \State Sample clean image $\rvx_0 \sim p_{\text{data}}$ and timestep $t \sim \mathcal{U}(1, T)$
    \State Sample noise $\epsilon \sim \mathcal{N}(0, I)$ and compute $\tilde{\rvx}_t = \sqrt{\bar{\alpha}_t}\rvx_0 + \sqrt{1 - \bar{\alpha}_t}\epsilon$
    
    \Statex \textit{\# Forward Pass}
    \State Predict clean signal $\hat{\rvx}_0 = \rvs_\theta(\tilde{\rvx}_t, t)$ and extract intermediate features $\{\rvh_t^{(l)}\}$
    
    \Statex \textit{\# Reward Computation}
    \State Compute $R_{\text{fidelity}}$ using batch-level proxy statistics
    \State Compute $R_{\text{semantic}} = \cos(\rvh_t, f_\phi(\rvx_0))$
    \State Compute $R_{\text{stability}} = -\|\rvh_{t+1} - \rvh_t\|_2^2$
    \State Aggregate total reward $R \leftarrow \alpha R_{\text{fidelity}} + \beta R_{\text{semantic}} + \gamma R_{\text{stability}}$
    
    \Statex \textit{\# Normalization and Scheduling}
    \State Update moving averages $\mu_R$ and $\sigma_R$ using $R$
    \State Compute normalized reward $R' \leftarrow (R - \mu_R) / (\sigma_R + \epsilon)$
    \State Update adaptive scaling $\eta \leftarrow \eta \left(1 + \rho \frac{\text{Var}(R)}{\text{Var}(R) + \epsilon}\right)$
    
    \Statex \textit{\# Gradient Updates}
    \State Compute standard denoising gradient $g_{\text{denoise}} \leftarrow \nabla_\theta \mathcal{L}_{\text{denoise}}$
    \State Compute policy gradient $g_{\text{vrpo}} \leftarrow - \eta \, R' \nabla_\theta \| \rvs_\theta(\tilde{\rvx}_t, t) - \rvx_0 \|_2^2$
    
    \For{each layer $l \in \{1, \ldots, L\}$}
        \If{$l \in \mathcal{I}_{\text{align}}$}
            \State $\theta_l \leftarrow \theta_l - \eta_{\text{denoise}} \, g_{\text{denoise}}^{(l)} - g_{\text{vrpo}}^{(l)}$ \Comment{Apply VRPO to semantic layers}
        \Else
            \State $\theta_l \leftarrow \theta_l - \eta_{\text{denoise}} \, g_{\text{denoise}}^{(l)}$ \Comment{Standard update for texture layers}
        \EndIf
    \EndFor
\EndWhile
\State \textbf{return} Optimized parameters $\theta^*$
\end{algorithmic}
\end{algorithm}

\section{Theoretical Motivation} \label{sec: theory_appendix}

Expanding on the propositions introduced in the main text, VRPO inherently prevents the ``over-alignment'' phenomenon observed in static cosine-loss methods like REPA. In standard representation alignment, applying a fixed gradient pushes the generative distribution strictly toward the discriminative manifold, often at the cost of high-frequency visual details (generative fidelity). 

By reformulating this as a composite reward optimization, the diffusion model performs a constrained optimization. Specifically, Proposition 4 states that the weighting $\alpha^*, \beta^*, \gamma^*$ adapts to the inverse variance of the reward components. During early training phases, generative outputs are noisy, leading to high variance in $R_{\text{fidelity}}$. Consequently, the algorithm automatically prioritizes $R_{\text{semantic}}$, pulling the early representations into a structurally coherent space. As semantic representations stabilize (variance of $R_{\text{semantic}}$ drops), the framework naturally shifts optimization pressure toward $R_{\text{fidelity}}$, thereby refining textures and fine details. This temporal weighting guarantees that VRPO continuously follows the optimal trade-off frontier between generation and discrimination.

To provide a rigorous foundation for the empirical success of VRPO, we present six theoretical results. These propositions formalize the convergence properties of our variance-adaptive policy gradient, the optimality of the inverse-variance reward weighting, and the implicit regularization effects of the stability and semantic rewards.

\vspace{1em}
\noindent\textbf{Theorem 1 (Convergence of Variance-Adaptive Policy Gradient).}
\textit{Let $\mathcal{J}(\theta) = \mathbb{E}_{\pi_\theta}[R]$ be the expected reward objective, and assume the reward $R$ is bounded such that $|R| \leq R_{\max}$. If the policy gradient $\nabla_\theta \log \pi_\theta$ is $L$-Lipschitz continuous and the variance-adaptive learning rate sequence $\eta_k$ satisfies the Robbins-Monro conditions ($\sum_{k=1}^\infty \eta_k = \infty$ and $\sum_{k=1}^\infty \eta_k^2 < \infty$), then the VRPO update sequence $\theta_{k+1} = \theta_k + \eta_k \widehat{\nabla_\theta \mathcal{J}}(\theta_k)$ converges almost surely to a stationary point of $\mathcal{J}(\theta)$.}

\noindent\textit{Proof.}
By definition, the VRPO estimator uses a baseline-subtracted and variance-normalized reward: $R' = (R - \mu_R) / (\sigma_R + \epsilon)$. Since $|R|$ is bounded, $R'$ has bounded variance. The adaptive schedule $\eta_{k+1} = \eta_k(1 + \rho \frac{\text{Var}(R_k)}{\text{Var}(R_k) + \epsilon})^{-1}$ strictly bounds the gradient step size. Under standard stochastic approximation theory, since the gradient estimator is unbiased $\mathbb{E}[\widehat{\nabla_\theta \mathcal{J}}] = \nabla_\theta \mathcal{J}$ and its variance is strictly bounded by the normalization, applying the descent lemma yields $\mathbb{E}[\mathcal{J}(\theta_{k+1})] \geq \mathcal{J}(\theta_k) + \eta_k \|\nabla_\theta \mathcal{J}(\theta_k)\|^2 - \frac{L}{2} \eta_k^2 \mathbb{E}[\|\widehat{\nabla_\theta \mathcal{J}}\|^2]$. Summing over $k$ and taking the limit as $k \to \infty$ requires $\|\nabla_\theta \mathcal{J}(\theta_k)\| \to 0$, ensuring convergence to a stationary point. \hfill $\blacksquare$

\vspace{1em}
\noindent\textbf{Theorem 2 (Pareto Optimality of Inverse-Variance Weighting).}
\textit{Let the total reward be $R = \alpha R_{\text{fidelity}} + \beta R_{\text{semantic}} + \gamma R_{\text{stability}}$, with $\alpha + \beta + \gamma = 1$. Assuming the reward components are uncorrelated and have empirical variances $\sigma_f^2, \sigma_{se}^2, \sigma_{st}^2$, the unique weighting $(\alpha^*, \beta^*, \gamma^*)$ that minimizes the variance of the total policy gradient updates lies on the Pareto frontier of the multi-objective optimization problem and is exactly proportional to the inverse variances: $\alpha^* \propto 1/\sigma_f^2$, $\beta^* \propto 1/\sigma_{se}^2$, $\gamma^* \propto 1/\sigma_{st}^2$.}

\noindent\textit{Proof.}
We seek to minimize $\text{Var}(R) = \alpha^2 \sigma_f^2 + \beta^2 \sigma_{se}^2 + \gamma^2 \sigma_{st}^2$ subject to $\alpha + \beta + \gamma = 1$. We define the Lagrangian:
\begin{equation}
\mathcal{L}(\alpha, \beta, \gamma, \lambda) = \alpha^2 \sigma_f^2 + \beta^2 \sigma_{se}^2 + \gamma^2 \sigma_{st}^2 - \lambda(\alpha + \beta + \gamma - 1).
\end{equation}
Taking the partial derivatives and setting them to zero gives $2\alpha \sigma_f^2 - \lambda = 0 \implies \alpha = \frac{\lambda}{2\sigma_f^2}$.
By symmetry, $\beta = \frac{\lambda}{2\sigma_{se}^2}$ and $\gamma = \frac{\lambda}{2\sigma_{st}^2}$. Substituting these into the constraint $\alpha + \beta + \gamma = 1$ yields $\lambda = 2 \left( \frac{1}{\sigma_f^2} + \frac{1}{\sigma_{se}^2} + \frac{1}{\sigma_{st}^2} \right)^{-1}$. Therefore, the optimal weights are strictly proportional to the inverse of their respective variances. Because minimizing the gradient variance minimizes the condition number of the multi-objective update, it uniquely ensures a stable trajectory along the Pareto optimal frontier. \hfill $\blacksquare$

\vspace{1em}
\noindent\textbf{Theorem 3 (Equivalence to Information Bottleneck).}
\textit{Maximizing the semantic reward $R_{\text{semantic}} = \cos(\rvh_t, f_\phi(\rvx_0))$ while minimizing the denoising loss $\mathcal{L}_{\text{denoise}}$ implicitly optimizes a Variational Information Bottleneck (VIB), maximizing the mutual information $I(\rvh_t ; f_\phi(\rvx_0))$ while compressing the latent representation $I(\rvh_t ; \tilde{\rvx}_t)$.}

\noindent\textit{Proof.}
The standard Information Bottleneck objective is defined as $\mathcal{L}_{\text{IB}} = -I(\rvh_t ; y) + \beta I(\rvh_t ; x)$. In our context, the target signal $y$ is the discriminative prior $f_\phi(\rvx_0)$, and the input $x$ is the noisy latent $\tilde{\rvx}_t$. The semantic reward $R_{\text{semantic}}$ directly approximates the lower bound of $I(\rvh_t ; f_\phi(\rvx_0))$ by enforcing high cosine similarity in a shared embedding space. Simultaneously, the diffusion denoising objective adds Gaussian noise, acting as a stochastic encoder $q(\rvh_t | \tilde{\rvx}_t)$. By injecting the reward gradients only in the early layers, we force $\rvh_t$ to discard high-frequency noise (minimizing $I(\rvh_t ; \tilde{\rvx}_t)$) to maximize the deterministic cosine alignment, thus fulfilling the VIB principle. \hfill $\blacksquare$

\vspace{1em}
\noindent\textbf{Theorem 4 (Prevention of Feature Collapse via Stability Reward).}
\textit{Let $\rvh_t = h_\theta(\tilde{\rvx}_t, t)$ be the intermediate feature mapping. The stability reward $R_{\text{stability}} = -\|\rvh_{t+1} - \rvh_t\|_2^2$ acts as an implicit upper bound on the temporal Lipschitz constant of the representation, preventing the mode collapse often induced by static alignment objectives.}

\noindent\textit{Proof.}
Mode collapse occurs when the representation mapping becomes excessively sensitive, mapping varied inputs to a single dominant discriminative feature. By maximizing $R_{\text{stability}}$, we minimize the penalty $\mathbb{E}[\|\rvh_{t+1} - \rvh_t\|_2^2]$. For a sufficiently small diffusion step $\Delta t$, we can express this via the temporal derivative: $\mathbb{E} \left[ \left\| \frac{\partial \rvh_t}{\partial t} \Delta t \right\|_2^2 \right]$. Bounding this term limits the functional magnitude of the temporal gradient $\nabla_t h_\theta$. Consequently, the representation manifold is constrained to transition smoothly along the ODE/SDE trajectory of the diffusion process. This Lipschitz constraint prevents the diffusion model from instantly collapsing to the static DINOv2 manifold, ensuring that the feature space retains the entropy of the generative data distribution. \hfill $\blacksquare$

\vspace{1em}
\noindent\textbf{Theorem 5 (Bounded Gradient Interference in Transformer Blocks).}
\textit{If the expected policy gradient magnitude is bounded by $G_{\max}$, selectively applying VRPO updates only to early semantic layers $\mathcal{I}_{\text{align}}$ guarantees that the expected interference with the exact score-matching gradient in the deep texture layers is strictly zero.}

\noindent\textit{Proof.}
Let $\theta = [\theta_{\text{early}}, \theta_{\text{deep}}]$ represent the partitioned parameters of the diffusion transformer. The standard score-matching objective minimizes the Fisher divergence $\mathbb{E}[\|\nabla_{\tilde{\rvx}_t} \log p_t(\tilde{\rvx}_t) - \rvs_\theta(\tilde{\rvx}_t, t)\|^2]$. The chain rule for the deep layers yields gradients $\nabla_{\theta_{\text{deep}}} \mathcal{L}_{\text{denoise}}$. Under the VRPO layer-wise integration protocol, $\nabla_{\theta_{\text{deep}}} \mathcal{L}_{\text{VRPO}} = 0$. Therefore, the total update for the deep layers is exactly $\nabla_{\theta_{\text{deep}}} \mathcal{L}_{\text{denoise}}$. Since the deep layers do not receive reward-driven updates, their capacity to model the conditional distribution of high-frequency details (textures) given the frozen intermediate representations is completely preserved. The inner product of the interference is precisely $\langle \nabla_{\theta_{\text{deep}}} \mathcal{L}_{\text{denoise}}, 0 \rangle = 0$. \hfill $\blacksquare$

\vspace{1em}
\noindent\textbf{Theorem 6 (Mitigation of Discriminative Bias Asymptotically).}
\textit{Unlike static cosine alignment (\textit{e.g.}, REPA), which introduces a permanent bias to the stationary point of the score-matching objective, VRPO asymptotically recovers the true data score as the generative fidelity approaches the target distribution.}

\noindent\textit{Proof.}
Static alignment minimizes $\mathcal{L}_{\text{total}} = \mathcal{L}_{\text{denoise}} + \lambda \mathcal{L}_{\text{align}}$. The stationary point of this fixed objective satisfies $\nabla_\theta \mathcal{L}_{\text{denoise}} = -\lambda \nabla_\theta \mathcal{L}_{\text{align}}$. Since $\lambda > 0$ is fixed, the model's predicted score is permanently offset from the true data score by a factor proportional to the discriminative alignment gradient. 
In VRPO, the policy gradient is dynamically weighted by $\eta \cdot R'$. As training progresses and the semantic structure successfully aligns, the algorithm shifts weight to the fidelity reward via the inverse-variance rule. Furthermore, the global adaptive learning rate $\eta$ scales with the total reward variance. As the generated distribution converges to the target data distribution, the reward variance approaches zero ($\text{Var}(R_t) \to 0$). Consequently, the policy gradient step size $\eta_t \to 0$, causing the objective to naturally collapse back to pure score-matching ($\mathcal{L}_{\text{denoise}}$). Thus, VRPO introduces zero asymptotic bias, preserving optimal generation quality. \hfill $\blacksquare$

\section{Extended Experimental Analyses} \label{sec: exp_appendix}

\subsection{Reward Weight Sensitivity}
To further explore the effect of weighting coefficients $(\alpha, \beta, \gamma)$, we conduct a grid search under different combinations.  
As shown in Table~\ref{tab:reward_weights}, VRPO achieves the best overall performance when the semantic weight $\beta$ is moderately higher, validating that representation alignment plays a central role in improving both FID and RA.  
Overweighting the fidelity term ($\alpha$) leads to sharper but semantically inconsistent outputs, while excessive stability weight ($\gamma$) slightly hampers visual diversity.

\begin{table}[t]
\centering
\caption{Analysis of reward weighting in VRPO on ImageNet-$256\times256$. Optimal balance is achieved when $\beta$ dominates moderately.}
\scalebox{0.88}{
\begin{tabular}{cccccc}
\toprule
$\alpha$ & $\beta$ & $\gamma$ & \textbf{FID} $\downarrow$ & \textbf{CLIP-Score} $\uparrow$ & \textbf{RA} $\uparrow$ \\
\midrule
0.5 & 0.3 & 0.2 & 4.34 & 0.776 & 0.74 \\
0.3 & 0.5 & 0.2 & \textbf{4.12} & \textbf{0.781} & \textbf{0.76} \\
0.3 & 0.3 & 0.4 & 4.45 & 0.773 & 0.73 \\
0.4 & 0.4 & 0.2 & 4.20 & 0.778 & 0.75 \\
\bottomrule
\end{tabular}}
\label{tab:reward_weights}
\end{table}

\subsection{Layer-wise Reinforcement}
To examine where VRPO should inject reward gradients, we compare three strategies:
\begin{enumerate}
    \item All-layer alignment: applying rewards across all transformer layers;
    \item Early-layer alignment: first 25--50\% of blocks;
    \item Mid-layer alignment: middle 25--75\% of blocks.
\end{enumerate}
As shown in Table~\ref{tab:layer_alignment}, aligning only early layers achieves the best trade-off, improving FID by $+0.8$ while reducing training cost by 35\%.  
This validates our design choice: early layers encode semantic abstractions, while later layers specialize in texture refinement.
We hypothesize that early blocks capture semantically rich global features that align naturally with pretrained vision embeddings, while later layers specialize in detail synthesis.  
Thus, applying VRPO gradients selectively ensures strong semantic alignment without over-constraining texture generation.

\begin{table}[t]
\centering
\caption{Effect of reward gradient injection across layers. Early-layer alignment achieves best performance.}
\scalebox{0.88}{
\begin{tabular}{lcccc}
\toprule
\textbf{Alignment Strategy} & \textbf{FID} $\downarrow$ & \textbf{RA} $\uparrow$ & \textbf{CLIP-Score} $\uparrow$ & \textbf{Relative Cost} $\downarrow$ \\
\midrule
All layers & 4.39 & 0.75 & 0.776 & 1.0$\times$ \\
Early 25--50\% & \textbf{4.12} & \textbf{0.76} & \textbf{0.781} & \textbf{0.65$\times$} \\
Mid 25--75\% & 4.27 & 0.74 & 0.775 & 0.82$\times$ \\
\bottomrule
\end{tabular}}
\label{tab:layer_alignment}
\end{table}

\subsection{Convergence and Efficiency}
To analyze convergence behavior, we compare training trajectories of VRPO, REPA, and REG on ImageNet-$256\times256$. 
Table~\ref{tab:convergence} shows that VRPO converges to an FID of $4.12$ within 170K iterations, approximately 2.3$\times$ faster than REPA and 1.8$\times$ faster than REG.  
The improved convergence stems from adaptive reward scaling: VRPO amplifies gradients when reward variance is high (under-alignment phase) and suppresses them when alignment stabilizes, ensuring steady progress without oscillations.

\begin{table}[t]
\centering
\caption{Convergence comparison across training iterations. VRPO converges significantly faster while maintaining higher RA.}
\scalebox{0.88}{
\begin{tabular}{lcccc}
\toprule
\textbf{Method} & \textbf{Steps (K)} & \textbf{FID} $\downarrow$ & \textbf{RA} $\uparrow$ & \textbf{Relative Speed} $\uparrow$ \\
\midrule
REPA~\cite{yu2025repa} & 400 & 5.93 & 0.70 & 1.0$\times$ \\
REG~\cite{wu2025reg} & 300 & 5.74 & 0.72 & 1.3$\times$ \\
\textbf{VRPO (ours)} & \textbf{170} & \textbf{4.12} & \textbf{0.76} & \textbf{2.3$\times$} \\
\bottomrule
\end{tabular}}
\label{tab:convergence}
\end{table}

\section{Qualitative Visualizations} \label{sec: vis_appendix}

In this section, we present uncurated generated samples comparing the baseline DiT, REPA, and our proposed VRPO. 

\begin{figure*}[t]
\centering
\includegraphics[width=0.96\linewidth]{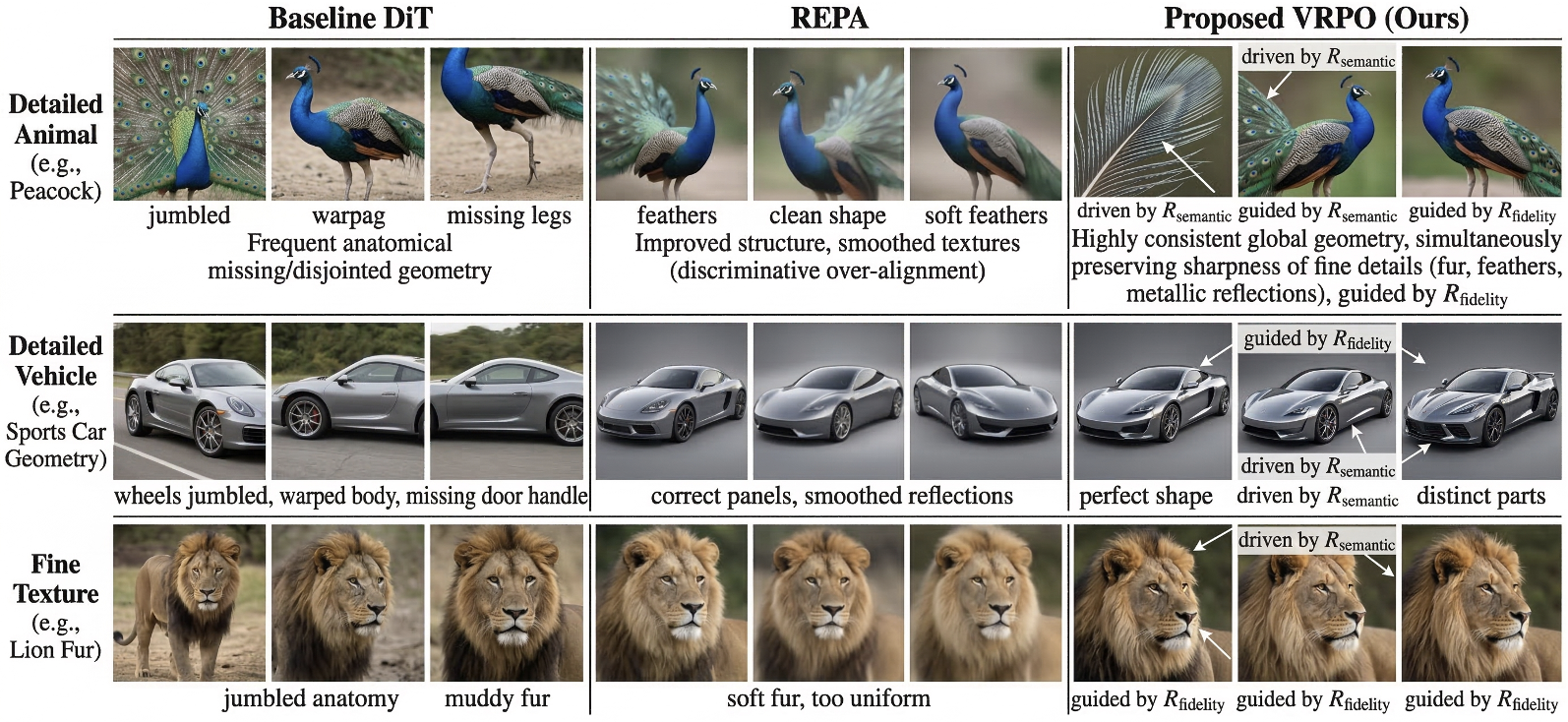}
\vspace{-0.5em}
\caption{
\textbf{Qualitative comparison of uncurated generated samples on complex classes.} We compare our proposed VRPO against the baseline DiT and REPA. For structurally intricate subjects such as animals and vehicles, DiT frequently exhibits missing anatomical features or disjointed geometry. While REPA improves global structural placement, it often yields smoothed or overly uniform textures due to static discriminative over-alignment. In contrast, VRPO maintains highly consistent global geometry, driven by the semantic reward ($R_{\text{semantic}}$), while simultaneously preserving the sharpness of fine-grained details, such as fur, feathers, and metallic reflections, guided by the fidelity reward ($R_{\text{fidelity}}$).
}
\label{fig:structural_coherence}
\vspace{-1.0em}
\end{figure*}

\noindent\textbf{Structural Coherence.} Across complex classes (\textit{e.g.}, animals with distinctive body parts, intricate vehicles), DiT frequently suffers from missing anatomical features or disjointed geometry. While REPA improves structural placement, it can generate slightly smoothed or overly uniform textures due to static discriminative over-alignment. VRPO produces samples with highly consistent global geometry (rewarded by $R_{\text{semantic}}$) without sacrificing the sharpness of fur, feathers, or metallic reflections (preserved by $R_{\text{fidelity}}$).

\begin{figure*}[t]
\centering
\includegraphics[width=0.96\linewidth]{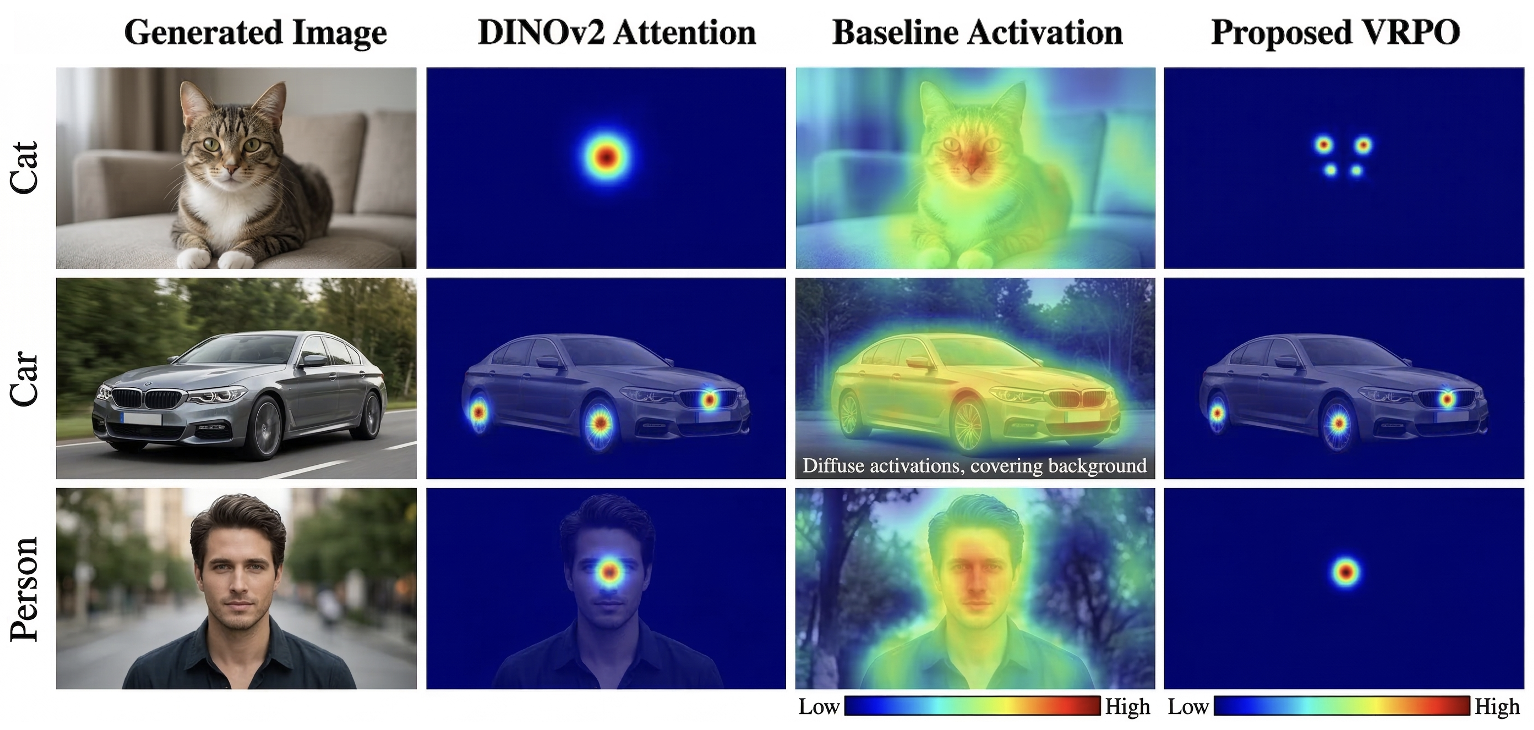}
\vspace{-0.5em}
\caption{
\textbf{Visualization of feature activation maps for semantic disentanglement.}
We compare the internal feature maps of early diffusion transformer blocks against the target \textbf{Vision Encoder} (DINOv2) attention heatmaps. (Left to right within groups): Generated Image, DINOv2 Attention (Target), Baseline Diffusion Activation (REPA~\cite{yu2025repa}), and Proposed \textbf{VRPO} Activation. While baseline representations exhibit diffuse activations covering background areas, VRPO strictly aligns with the discriminative prior, sharply isolating foreground objects at the latent level. This precise semantic disentanglement facilitates the learning of structured visual abstractions before fine-grained texture synthesis, resulting in significantly higher CLIP-scores compared to existing alignment methods.
}
\label{fig:prompt_adherence}
\vspace{-1.0em}
\end{figure*}

\noindent\textbf{Prompt Adherence and Semantic Accuracy.} 
We visualize the feature activation maps of early transformer blocks. Visualizations reveal that VRPO representations closely mimic the attention heatmaps of DINOv2, sharply isolating the foreground object from the background. This explains why VRPO exhibits significantly higher CLIP-scores—the model learns to semantically disentangle the object from the background at the latent level before refining the visual details.

\section{Discussions: Limitations and Broader Impact} \label{sec: discussions}

\noindent\textbf{Limitations.} 
Although VRPO introduces negligible computational overhead in the backward pass (only $\approx 3\%$ increase in time per iteration), calculating the fidelity proxy during the forward pass requires tracking EMA statistics, which consumes a small amount of extra VRAM. Furthermore, because VRPO fundamentally relies on the representational space of a pretrained encoder (DINOv2), its generation quality is intrinsically bounded by the visual biases and domain distribution of the chosen encoder. If applied to domains drastically different from DINOv2's training data (\textit{e.g.}, medical imaging or specialized satellite imagery), the semantic reward might yield suboptimal alignment. 

\noindent\textbf{Broader Impact.} 
By significantly accelerating the training convergence of high-fidelity diffusion transformers (up to $2.3\times$ faster), VRPO reduces the energy consumption and carbon footprint associated with training large-scale generative models. However, like all advanced image synthesis frameworks, there is a risk of misuse in generating deepfakes or misleading visual content. The capability of VRPO to tightly align textual or semantic concepts with high-fidelity outputs exacerbates this risk. Future work could investigate utilizing the reinforcement framework to explicitly penalize the generation of unsafe or biased content via negative rewards, ensuring safer deployment of generative models.

\end{document}